\documentclass{article}
\usepackage{hyperref}
\usepackage{url}
\usepackage{lineno,hyperref}
\usepackage{graphicx}
\usepackage{amsmath}
\usepackage{listings}
\usepackage{bm}
\usepackage{adjustbox}
\usepackage{diagbox}
\usepackage{rotating}
\usepackage{multirow}
\usepackage{commath}
\usepackage{amssymb}
\usepackage{geometry}
\modulolinenumbers[5]

\title{For2For: Learning to forecast from forecasts}

\author{
  Zhao, Shi\\
  \normalsize{Tencent} \\
  \texttt{shi.zhao@foxmail.com}
  \and
  Feng, Ying\\
  \normalsize {IBSS, XJTLU} \\
  \texttt{ying.feng@xjtlu.edu.cn}
}

\begin{document}
\maketitle
\begin{abstract}
This paper presents a time series forecasting framework which combines standard forecasting methods and a machine learning model. The inputs to the machine learning model are not lagged values or regular time series features, but instead forecasts produced by standard methods. The machine learning model can be either a convolutional neural network model or a recurrent neural network model.  The intuition behind this approach is that forecasts of a time series are themselves good features characterizing the series, especially when the modelling purpose is forecasting. It can also be viewed as a weighted ensemble method. Tested on the M4 competition dataset, this approach outperforms all submissions for quarterly series, and
is more accurate than all but the winning algorithm for monthly series.
\end{abstract}
%\linenumbers
\section{Introduction}
The competitiveness of neural network (NN) models and other machine learning (ML) models for time series forecasting compared to statistical models has long been questioned by practitioners \cite{makridakis2018statistical} \cite{makridakis2019forecasting}. Although in the field of time series forecasting, there is a plethora of literature presenting complex novel models,  in practice the performance of ML models is often below expectation \cite{hewamalage2019recurrent}. For example, pure ML methods perform poorly in the M4 competition, with none of the five submitted pure ML solutions beating the Comb benchmark, which is the simple arithmetic average of Single, Holt and Damped exponential smoothing \cite{makridakis2018m4}.

The poor performance of NN and general ML models in forecasting time series is in sharp contrast with the tremendous success of these models in the areas like computer vision and natural language processing. NN models are known to be capable of extracting features automatically from images for tasks like classification, while in the pre-deep learning era, such tasks were usually solved by feeding hand-crafted features to various ML models \cite{nanni2017handcrafted}. The availability of large dataset such as ImageNet, which has more than 14 million images, is believed to be crucial for the success of NN models in computer vision.

The lack of enough time series data is attributed as a reason for the underachievement of ML models in forecasting \cite{hewamalage2019recurrent}. Although it is shown by Zhang \emph{et al.} \cite{rong2018forward} that there exists feedforward NNs that are able to generate many popular time series features such as minimum, maximum and count above mean,
without enough data, sophisticated models easily get overfitted and fail short in extracting true patterns. When the search space is so large and the guidance is so little, it is not surprising that ML models are not lucky enough to find the parameters to generalize well.

To make ML models work for time series forecasting, we need to provide more guidance, which can be either more data or better features. But usually it may be difficult or even impractical to collect enough relevant series. Therefore, generating good features seems to be a more viable option. After all, before deep learning gained popularity, the key to the success of many ML systems is feature engineering. Some even believe that ``applied machine learning is basically feature engineering" \cite{ng2013machine}.

%In this paper, we present two neural network models, one convolutional and one recurrent, for time series forecasting. The inputs to these models are not lagged values or extracted features like seasonality strength, but instead the forecasts produced by standard off-the-shelf forecasting methods such as ARIMA and ETS. The intuition behind this choice is that the forecasts of a time series are themselves very good features characterizing the series, especially when the modelling purpose is forecasting. In other words, the guidance we provide to the NN models are the forecasts produced by other models, which we will refer to as base models.

This paper presents a time series forecasting framework we call \emph{For2For} (forecasts to forecast). The framework is composed of standard off-the-shelf forecasting methods and a ML model. The inputs to the ML model are not lagged values or mined features like seasonality strength, but instead forecasts produced by standard methods such as ARIMA and ETS.  The intuition behind this approach is that forecasts of a time series are themselves good features characterizing the series, especially when the modelling purpose is forecasting. The ML model can be either a convolutional  neural network (CNN) model or a recurrent neural network (RNN) model. In other words, the guidance we provide to the NN models are the forecasts produced by other models, which we will refer to as base models.

Of course, a different way of viewing this approach is that the NN models are trained to combine the forecasts of base models, therefore essentially it is an ensemble model. But in our opinion, it is different from standard combination methods in two ways. First of all, this method does not try to select a single base model (as Talagala \emph{et al.}
%, Hyndman \& Athanasopoulos
\cite{talagala2018meta}) or to assign weights to base models (as FFORMA \cite{montero2020fforma}). Instead it tries to combine the forecasts of base models differently at each forecasting point. We will elaborate this further in Section 2. Second, the inputs are forecasts of base models instead of time series features extracted by software packages such as \emph{tsfeatures} \cite{hyndman2019tsfeatures}.

This method is tested on the M4 competition dataset and it is more accurate than a simple arithmetic combination of the base models. Furthermore, it beats all submissions for quarterly series, and outperforms all but the winning algorithm for monthly series.
%it outperforms all but the top two submissions in the competition for yearly, quarterly and monthly series. The accuracy of this approach is very close to FFORMA method.

The rest of the paper is organized as follows. Section 2 introduces the framework and presents the structures of the CNN model and the RNN model. The implementation details and testing results are discussed in Section 3 and Section 4 concludes the paper.

\section{Methodology}
Let us first define $\bm{X}=\{X_1,X_2,...,X_N\}$ as a collection of time series to forecast, and the forecasting horizon is set to $h$.  For notational convenience, the first time step to forecast is defined as $t=1$, therefore $X_n$ is known up to $t=0$ with $n={1,2,...,N}$.

\subsection{The For2For framework}
\begin{figure}[bhp]
\centering
\includegraphics[width=80mm]{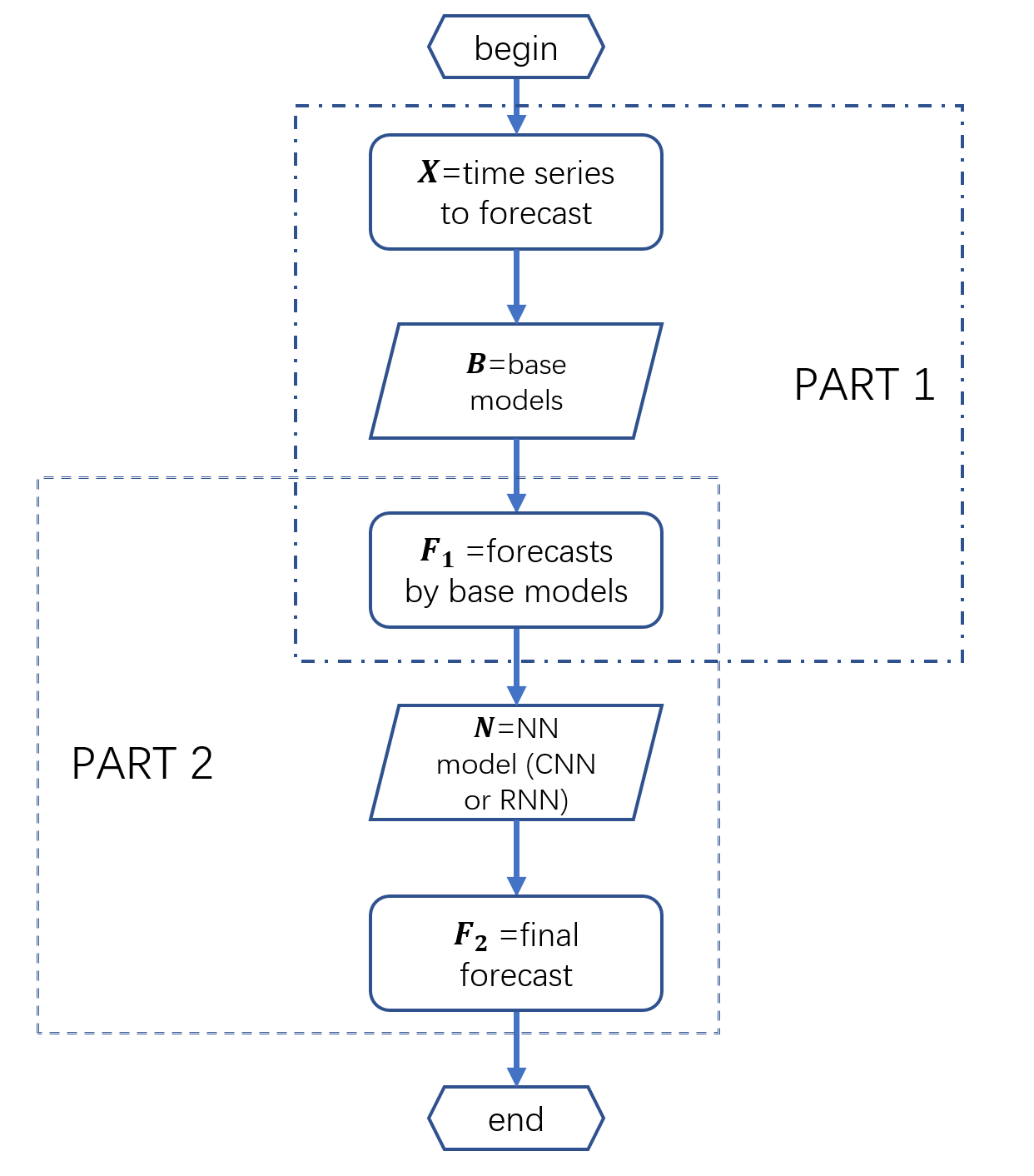}
\caption{For2For (forecasts to forecast) framework}
\label{For2For}
\end{figure}

%The two models presented in this paper fit well into the same framework which we call For2For (forecasts to forecast).
As illustrated in Figure \ref{For2For}, the \emph{For2For} framework consists of two parts. In the first part, a group of base models $\bm{B}=\{B_1,B_2,...,B_M\}$ are used to forecast time series $\bm{X}$, and the forecast results are
\begin{equation*}
\bm{F_1}=\{\hat{X}_{1,{B_1}},\hat{X}_{1,{B_2}},..,\hat{X}_{1,{B_M}};...;\hat{X}_{N,{B_1}},\hat{X}_{N,{B_2}},..,\hat{X}_{N,{B_M}}\}
\end{equation*}
where $\hat{X}_{n,{B_m}}$ denotes the forecast of $X_n$ produced by base model $B_m$. This is common practice in various forecast combination methods \cite{hyndman2018forecasting} \cite{montero2020fforma} \cite{pawlikowski2020weighted}. In the second part of the framework, the forecasts $\bm{F_1}$ produced by base models $\bm{B}$ are fed into a NN model $\bm{N}$. The NN model can be either a CNN model or a RNN model. %, with straightforward network structures.
The forecast $\bm{F_2}=\{\bar{X}_1,\bar{X}_2,..,\bar{X}_N\}$ generated by the NN model $\bm{N}$ is the final forecast result. Standard training and validation processes are used to build the NN model and are not drawn explicitly in the diagram.
%The main reasoning

In conventional combination approaches, the final forecast is a linear combination of forecasts generated by base models,
\begin{equation}\label{w1}
  \bar{X}_n = w_{n,B_1} \hat{X}_{n,{B_1}}+ w_{n,B_2} \hat{X}_{n,{B_2}}+ ... +w_{n,B_M} \hat{X}_{n,{B_M}}
\end{equation}
where $w_{n,B_m}$ is the weight assigned to base model $B_m$ when forecasting the series $X_n$.

In this framework, we do not try to select the single best performing base model, nor do we try to assign weights to each base model. In fact, no weights are explicitly calculated at all.
The NN model is trained to learn from the forecasts of base models and then make forecast automatically.

%If the NN model $\bm{N}$ is reduced to a linear model
%The weights are assigned to individual forecast
If we insist to view this approach as a weighted combination, then the weights are assigned to individual forecast result at each forecasting step,
\begin{equation}\label{w2}
  \bar{X}_n(t) = w_{n,B_1}(t) \hat{X}_{n,{B_1}}(t)+ w_{n,B_2}(t) \hat{X}_{n,{B_2}}(t)+ ... +w_{n,B_M}(t) \hat{X}_{n,{B_M}}(t)
\end{equation}
This means that the forecasts of a given series $X_n$ by a base model $B_m$  at different $t$ are not necessarily trusted to the same extent in a multi-horizon forecasting task. In other words, the weight $w$ is not just a function of $n$ (which denotes time series) and $m$ (which denotes base model), but also a function of $t$ (forecasting step).
It is only when $h=1$ is Equation \ref{w2} reduced to Equation \ref{w1}.

%\subsection{Base models}

Any popular forecasting methods can be included into the set of base models. After applying the base models to a time series $X_n$, we obtain a matrix $\hat{X}_n \in \mathbb{R}^{h\times M}$, with the $(i,j)$ element of the matrix being the forecast result at time $t=i$ by model $B_j$. Other relevant information such as the domain type in the M4 dataset can be appended side by side to this matrix.

\subsection{CNN model}\label{sec:cnn}
%The struture of the CNN model considered in this paper is shown in Figure \ref{cnn}, where in some sense the forecasts $\hat{X}_n$ is treated as an image with only one channel. The structure mimics the highly successful ResNet for image recognition \cite{he2016deep}.

\begin{figure}[bhp]
\centering
\includegraphics[width=80mm]{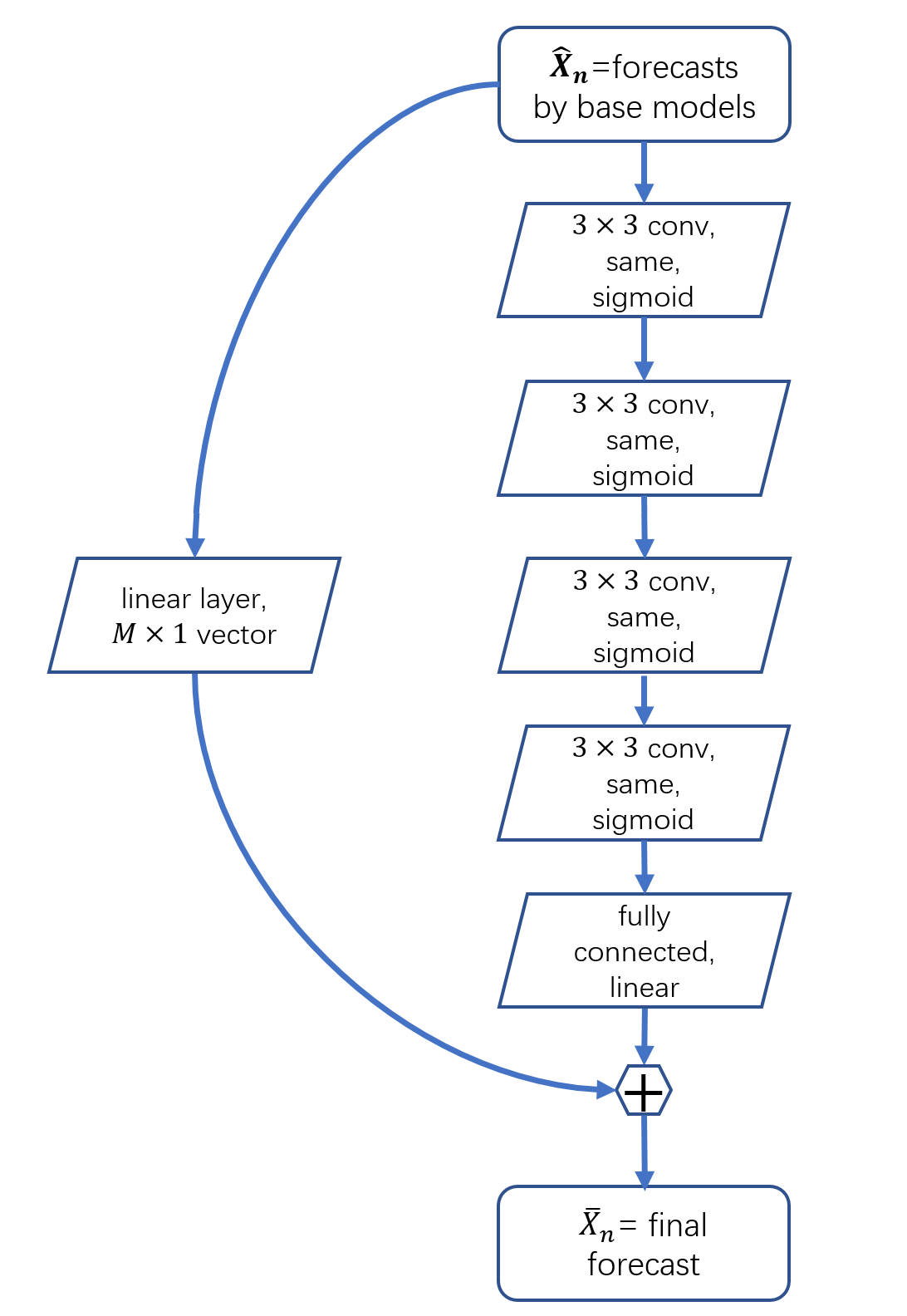}
\caption{CNN model structure}
\label{cnn}
\end{figure}

As shown in Figure \ref{cnn}, the CNN model considered in this paper mimics the highly successful ResNet for image recognition \cite{he2016deep}. In some sense the forecasts $\hat{X}_n$ is treated as an image with only one channel.

The linear layer with a $M \times 1$ parameter vector on the left simply combines the $M$ base models linearly, and the output is a $h \times 1$ vector.
This layer can also be viewed as a $1 \times M$ convolutional layer with $valid$ padding and linear activation. Often such a linear model is not adequate, then it is necessary to include more nonlinear mechanisms in the model, and this is exactly what the residual network on the right is intended to do.
Intuitively, the four $3 \times 3$ convolutional layers with $same$ padding and sigmoid activation are used to capture the difference between forecasts by different base models at different time steps. A fully connected layer which outputs a $h \times 1$ vector is connected to the last $3 \times 3$  convolutional layer. The number of $3 \times 3$ convolutional layers can be increased or decreased to reduce bias or variance when necessary. The fully connected layer could also be replaced by a $M\times 1$ linear layer if there is an obvious overfitting problem.
In the extreme case, all the $3 \times 3$ convolutional layers and the fully connected layer are removed and the model collapses to a linear combination model.

When there are extra categorical features available, they may be incorporated to the model by a separate embedding layer and the final forecast is the summation of the outputs from the linear part, the residual part and the embedding part.

%The right part of the scheme shows how to incorporate extra categorical information that might be available with an embedding layer. In the case of forecasting M4 dataset, $C$ in the embedding layer denotes the number of domain type and $h$ is again the forecasting horizon. It is clear that the right part is also just a simple linear mapping. The final forecast is the summation of the outputs from these three parts.

The CNN model is not flexible enough as to build the computation graph, it is necessary to know the forecasting horizon $h$ beforehand. This is a constraint placed by the fully connected layer. For this reason, when the forecasting horizons are different, as is the case for the series of different frequencies in the M4 competition, one model has to be built for each frequency.

\subsection{RNN model}
\begin{figure}[bhp]
\centering
\includegraphics[width=40mm]{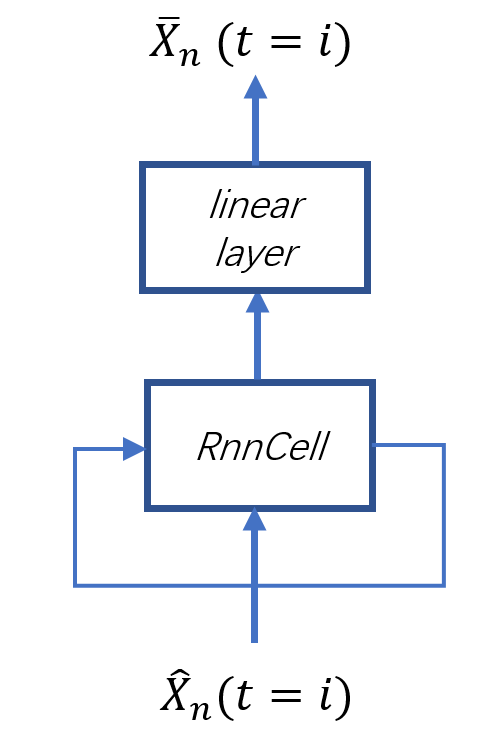}
\caption{RNN model structure, folded version}
\label{rnn_folded}
\end{figure}

\begin{figure}[bhp]
\centering
\includegraphics[width=\columnwidth]{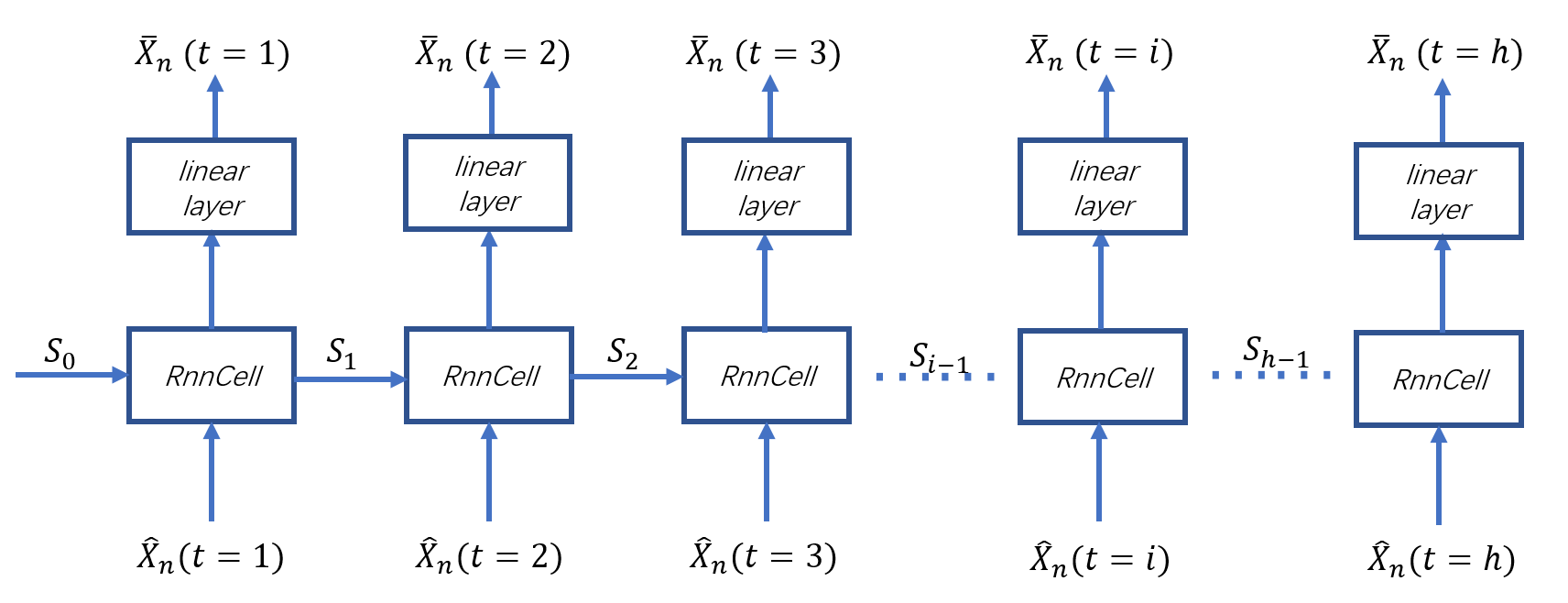}
\caption{RNN model structure, unfolded version}
\label{rnn_unfolded}
\end{figure}

The RNN model considered is more flexible than its CNN counterpart and is more interpretable. Figure \ref{rnn_folded} demonstrates the folded version of the RNN model considered in this paper, while Figure \ref{rnn_unfolded} shows the unfolded version, where $\hat{X}_n(t=i) \in \mathbb{R}^{M}$ is the vector containing the forecasts of series $X_n$ by all the $M$ base models at $t=i$, $S_{i-1}$ is the state vector of the RNN cell at $t=i-1$, $\bar{X}_n(t=i)$ is the final forecast of series $X_n$ at $t=i$. The initial state $S_0$ is set to zero. The linear layer is used to convert the output vector from the RNN cell to a scalar. In such a structure, instead of making all the $h$ forecasts at the very last step, the forecast is produced one at a time.

Due to the existence of $S_{i-1}$, $\bar{X}_n(t=i)$ is a combination of the forecasts $\hat{X}_n$ up to $t=i$. For example, when the model generates the final forecast at $t=2$, it does not only look at the forecasts by base models at $t=2$, which is $\hat{X}_n(t=2)$,  but also $\hat{X}_n(t=1)$. In such a model, it is not necessary to know the forecasting horizon $h$ in advance. It is also possible to make predictions for a different $h$ other than the one used for model training.

Extra information available for forecasting can be fed to the model by expanding the input vector to the RNN cell. For example, the domain type of the series in the M4 dataset can be included into the model by first applying an embedding layer to the categorical feature and then appending the embedding results to the forecasts by base models at each time step.

\section{Implementation and results}
 To assess the performance of the proposed method, the M4 dataset is used for experiments.
Among the 100,000 time series in the dataset, there are 23,000 yearly series, 24,000 quarterly series, 48,000 monthly series, 359 weekly series, 4227 daily series and 414 hourly series \cite{makridakis2018m4}. The forecasting horizons of series of these frequencies are six, eight, 18, 13, 14 and 48, respectively. The series were collected from six domains: micro, industry, macro, finance, demographic and other.

\subsection{Base models}
Inspired by FFORMA,  we choose the same base models as Montero-Manso \emph{et al.} \cite{montero2020fforma}, with the exception of the na\"ive method. The reason that we exclude the na\"{i}ve method is that at the data preprocessing stage, we normalize all the series by their last observations and then apply a log transformation. As a result, the forecasts at any time $t$ by the na\"ive method is simply zero. Including an all-zero vector into the inputs to a NN model adds no new information. % as it is equivalent to adding a bias term in a certain step.

For the sake of completeness, the eight base models are listed in Table \ref{bm_list}. These methods are implemented in the R package \emph{forecast} \cite{hyndman2007automatic} and can be readily used. We follow the practice in FFORMA so that the default settings of the R functions are used and the forecast is replaced by seasonal na\"ive forecast if an error is reported. %\table
\begin{table}
\centering
\begin{tabular}{ |p{5cm}|p{5cm}|}
 \hline
 base model & R function\\
 \hline
random walk with drift   & $rwf$ with $drift=TRUE$    \\
seasonal na\"ive &   $snaive$  \\
theta method & $thetaf$ \\
ARIMA model  & $auto.arima$ \\
exponential smoothing method & $ets$ \\
TBATS model& $tbats$  \\
STLM-AR & $stlm$ with $modelfunction=ar$ \\
feedforward neural network & $nnetar$ \\
 \hline
\end{tabular}
\caption{List of base models and R functions}
\label{bm_list}
\end{table}

\subsection{Preprocessing and training }
%The series are grouped by frequency
%As independent models are built for series of each frequencies, the series are grouped by frequency.
We take the monthly series as an example to show how the modelling process is carried out. First of all, the last 18 (horizon-number) observations of each of the 48,000 quarterly series are removed. Secondly, base models are applied to each chopped series to make forecasts. Negative forecasts are clipped by 10 as it is pointed out in \cite{makridakis2018m4} that the minimum of all series is 10. The clipped forecasts are normalized by the last observation and then a log transformation is applied.

At the model training stage, one third of randomly chosen series are held out as validation dataset for tuning model hyperparameters. Once the hyperparameters are determined, all the chopped series are used to re-train the final model, which will be used for the actual forecasting. The inputs to the final model are preprocessed forecasts of complete series produced by base models.
%At the actual forecasting stage, base models are applied to complete series (not chopped ones),

For fair comparison with FFORMA, the domain type of the series is not used in the experiments. Due to the stochastic nature of the initialization of model parameters and the fact that the training samples are randomly divided into $10$ mini-batches in each training epoch, the final model obtained is not deterministic. To reduce uncertainty, multiple instances of the model are trained concurrently and the average of these instances is taken as the final forecasting result. This model averaging approach was also employed by Smyl in his submission that wins the competition \cite{smyl2020hybrid}.

\subsection{Accuracy measure}
Mean absolute error (MAE) is used as the loss function. It is possible to improve the accuracy by using carefully designed custom loss function such as the pinball loss in \cite{smyl2020hybrid}. In the M4 competition, the accuracy of a model is measured by the symmetric mean absolute percentage error (sMAPE)
\begin{equation}\label{smape}
\text{sMAPE} = \frac{2}{h} \sum_{t=1}^{h} \frac{\abs{X(t)-\bar{X}(t)}}{\abs{X(t)}+\abs{\bar{X}(t)}}\times 100 (\%)
\end{equation}
and the mean absolute scaled error (MASE)
\begin{equation}\label{mase}
\text{MASE} = \frac{1}{h} \frac{\sum_{t=1}^{h}\abs{X(t)-\bar{X}(t)}}{\frac{1}{l-p}\sum_{t=p+1-l}^{0}\abs{X(t)-X(t-p)}}
\end{equation}
where $X(t)$ is the true value at point $t$, $\bar{X}(t)$ the forecast, $h$ the forecasting horizon, $l$ the number of in-sample data points and $p$ the time interval to compute the difference between successive data points \cite{makridakis2018m4}.

The final ranking of a submission in the competition is determined by the overall weight average (OWA) \cite{makridakis2018m4}, which is defined by
%\begin{equation}%\label{}
%  \text{OWA}_{\text{submission}} = \frac{1}{2}(\frac{\text{sMAPE}_{\text{submission}}}{\text{sMAPE}_{\text{Na\"ive2}}}+\frac{\text{MASE}_{\text{submission}}}{\text{MASE}_{\text{Na\"ive2}}})
%\end{equation}
\begin{equation}%\label{}
  \text{OWA} = \frac{1}{2}(\frac{\text{sMAPE}}{\text{sMAPE}_{\text{Na\"ive2}}}+\frac{\text{MASE}}{\text{MASE}_{\text{Na\"ive2}}})
\end{equation}
%where the subscript $_{\text{submission}}$ and $_{\text{Na\"ive2}}$ specify the
where the subscript $_{\text{Na\"ive2}}$ denotes the Na\"ive 2 method, which amounts to the na\"ive method applied to seasonally adjusted series. The Na\"ive 2 method
was used in the competition as the benchmark to evaluate the performance of submitted results \cite{makridakis2018m4}.

For ease of comparison, Table \ref{ranking} lists the accuracy measures of the top ranked submissions for series of each frequency in the M4 competition. Since the proposed method has much in common with FFORMA, the performance of the FFORMA method is given in the last column.
\begin{table}
\centering
%\begin{sidewaystable}
\begin{tabular}{|c|c|c|c|c|c|c|c||c|}\hline
%error measure & \multicolumn{3}{|c|}{sMAPE} \\ \hline
  measure &\diagbox{\small{frequency}}{\small{rank}}  & 1 & 2 & 3 & 4 & 5 & 6 & FFORMA \\ \hline
 \multirow{7}{*}{sMAPE} & Yearly & 13.176 & 13.366 & 13.528 & 13.669 & 13.673 & 13.677 & 13.528  \\ \cline{2-9}
 & Quarterly & 9.679 & 9.733 & 9.796 & 9.800 & 9.809 & 9.816 & 9.733 \\ \cline{2-9}
 & Monthly & 12.126 & 12.487 & 12.639 & 12.737 & 12.747 & 12.770 & 12.639 \\ \cline{2-9}
& Weekly & 6.582 & 6.726 & 6.728 & 6.814 & 6.905 & 6.919 & 7.625 \\ \cline{2-9}
& Daily & 2.452 & 2.852 & 2.959 & 2.980 & 2.985 & 2.993 & 3.097 \\ \cline{2-9}
& Hourly & 8.913 & 9.328 & 9.611 & 9.765 & 9.934 & 11.336 & 11.506 \\ \cline{2-9}
& Total & 11.374 & 11.695 & 11.720 & 11.836 & 11.845 & 11.887 & 	11.720	\\ \hline
 \multirow{7}{*}{MASE} & Yearly & 2.98 & 3.009 & 3.038 & 3.046 & 3.06 & 3.075 & 3.060  \\ \cline{2-9}
 & Quarterly & 1.111 & 1.118 & 1.118 & 1.122 & 1.125 & 1.134 & 1.111 \\ \cline{2-9}
 & Monthly & 0.884 & 0.893 & 0.895 & 0.905 & 0.907 & 0.913 & 0.893 \\ \cline{2-9}
& Weekly & 2.107 & 2.108 & 2.133 & 2.158 & 2.180 & 2.213 & 2.108 \\ \cline{2-9}
& Daily & 2.642 & 3.025 & 3.194 & 3.200 & 3.203 & 3.223 & 3.344	\\ \cline{2-9}
& Hourly & 0.801 & 0.810 & 0.810 & 0.819 & 0.856 & 0.861 & 0.819 \\ \cline{2-9}
& Total & 1.536 & 1.547 & 1.551 & 1.554 & 1.565 & 1.571 & 	1.551\\ \hline
 \multirow{7}{*}{OWA}  & Yearly & 0.778 & 0.788 & 0.799 & 0.801 & 0.802 & 0.805 & 0.799 \\ \cline{2-9}
 & Quarterly & 0.847 & 0.847 & 0.853 & 0.855 & 0.855 & 0.859 & 0.847 \\ \cline{2-9}
 & Monthly & 0.836 & 0.854 & 0.858 & 0.867 & 0.868 & 0.876 & 0.858	\\ \cline{2-9}
 & Weekly & 0.739 & 0.751 & 0.766 & 0.775 & 0.779 & 0.782 & 0.796 \\ \cline{2-9}
& Daily & 0.806 & 0.930 & 0.977 & 0.978 & 0.984 & 0.985 & 1.019	\\ \cline{2-9}
& Hourly & 0.410 & 0.440 & 0.444 & 0.474 & 0.477 & 0.477 & 0.484\\ \cline{2-9}
& Total & 0.821 & 0.838 & 0.841 & 0.842 & 0.843 & 0.848 & 	0.838 \\ \hline
\end{tabular}
\caption{Performance of top submissions}
\label{ranking}
%\end{sidewaystable}
\end{table}

%3.097	

\subsection{One model for each frequency}
%Since model parameters are initialized randomly and in the training process, the training samples are divided into $10$ batches randomly in each epoch, the results of the model
In this subsection, independent models are built for series of different frequencies.
In the CNN model, four layers of $3 \times 3$ convolutional layers are used for yearly, quarterly, and monthly series.
LSTM (long short-term memory) networks are used in the RNN model and the numbers of cell states are three for yearly series, four for quarterly series and six for monthly series.

We train each model eight times with fresh starts and in each run, the model is trained 2000 epochs. The sMAPE, MASE and OWA of each run for each frequency are shown in Table \ref{owa}. The last column of the table gives the accuracy of the ensemble model. It is clear that model averaging improves the forecasting accuracy.  In some cases, the emsemble is more accurate than any of the eight individual runs.

The proposed CNN model would rank 3rd, 2nd and 3rd in terms of sMAPE for yearly, quarterly and monthly series. It would rank 6th, 1st and 3rd in terms of OWA, respectively.
From Table \ref{owa} it can be seen that the RNN model is more accurate than its CNN counterpart in terms of sMAPE. The benefits of model averaging is even more evident. For yearly, quarterly and monthly series, the RNN model would rank 3rd, 1st and 2nd, respectively, in terms of sMAPE, and would rank 8th, 1st and 2nd, respectively, in terms of OWA. %Overall, the performance of this method is very close to that of FFORMA for these frequencies.

The CNN and the RNN models for yearly series have very good rankings in terms of sMAPE, but not so in terms of MASE. This is due to the difference of the two measures. The numerators in Equation \ref{smape} and Equation \ref{mase} are the same, while the denominators are different. For series with large mean value but very small variations, it is possible that the sMAPE of a forecast is very small, but the MASE is excessively high.

In our experiments, forecasts of base models are log transformed after being normalized by the last in-sample observation. Such a combination of preprocessing is well suited for reducing sMAPE. But when the purpose is to minimise MASE, it may be better not to do log transformation, but to simply normalize the forecasts of base models by the denominator of Equation \ref{mase}. To strike a balance between these two measures, it is possible to combine forecasts under these two preprocessing settings. We do not do so in this paper, as our purpose is to show the generality of the proposed framework for time series forecasting. It should be mentioned that there are many possible ways to improve the implementation.

\begin{sidewaystable}
\centering
\begin{tabular}{|c|c|c|c|c|c|c|c|c|c|c||c|}\hline
%error measure & \multicolumn{3}{|c|}{sMAPE} \\ \hline
model & measure &\diagbox{frequency}{instance}  & 1 & 2 & 3 & 4 & 5 & 6 & 7 & 8 & emsemble \\ \hline
\multirow{9}{*}{CNN} & \multirow{3}{*}{sMAPE} & Yearly & 13.5649 & 13.4904 & 13.5597 & 13.5856 & 13.5773 & 13.5090 & 13.5964 & 13.5848 & 13.4984 \\ \cline{3-12}
& & Quarterly & 9.6966 & 9.7330 & 9.7050 & 9.6943 & 9.6690 & 9.7116 & 9.6926 & 9.6933 & 9.6802 \\ \cline{3-12}
& & Monthly & 12.5993 & 12.6282 & 12.6066 & 12.6343 & 12.5937 & 12.7182 & 12.5805 & 12.6811 & 12.5478 \\ \cline{2-12}
& \multirow{3}{*}{MASE} & Yearly & 3.1325 & 3.1023 & 3.1325 & 3.1587 & 3.1427 & 3.1096 & 3.1309 & 3.1441 & 3.1150 \\ \cline{3-12}
& & Quarterly & 1.1060 & 1.1113 & 1.1067 & 1.1055 & 1.1033 & 1.1098 & 1.1040 & 1.1087 & 1.1038 \\ \cline{3-12}
& & Monthly & 0.9037 & 0.9024 & 0.9003 & 0.9034 & 0.9039 & 0.9088 & 0.9061 & 0.9089 & 0.8964 \\ \cline{2-12}
& \multirow{3}{*}{OWA}  & Yearly & 0.8092 & 0.8031 & 0.8090 & 0.8131 & 0.8108 & 0.8046 & 0.8099 & 0.8112 & 0.8049 \\ \cline{3-12}
& & Quarterly & 0.8436 & 0.8472 & 0.8443 & 0.8434 & 0.8414 & 0.8457 & 0.8427 & 0.8445 & 0.8421 \\ \cline{3-12}
& & Monthly & 0.8617 & 0.8621 & 0.8604 & 0.8628 & 0.8616 & 0.8682 & 0.8622 & 0.8670 & 0.8565 \\ \hline
\multirow{9}{*}{RNN} & \multirow{3}{*}{sMAPE} & Yearly & 13.5601 & 13.5879 & 13.5949 & 13.5696 & 13.5649 & 13.5490 & 13.5549 & 13.5580 & 13.4928 \\ \cline{3-12}
& & Quarterly & 9.6981 & 9.7016 & 9.7020 & 9.6991 & 9.7048 & 9.7034 & 9.7054 & 9.6825 & 9.6610 \\ \cline{3-12}
& & Monthly & 12.5217 & 12.5556 & 12.5922 & 12.5097 & 12.5527 & 12.5240 & 12.5028 & 12.5702 & 12.4770 \\ \cline{2-12}
& \multirow{3}{*}{MASE} & Yearly & 3.1415 & 3.1498 & 3.1470 & 3.1463 & 3.1357 & 3.1416 & 3.1309 & 3.1424 & 3.1254 \\ \cline{3-12}
& & Quarterly & 1.1077 & 1.1068 & 1.1076 & 1.1088 & 1.1082 & 1.1076 & 1.1088 & 1.1043 & 1.1051 \\ \cline{3-12}
& & Monthly & 0.8927 & 0.8930 & 0.8965 & 0.8926 & 0.8931 & 0.8949 & 0.8912 & 0.8942 & 0.8895 \\ \cline{2-12}
& \multirow{3}{*}{OWA}  & Yearly & 0.8101 & 0.8120 & 0.8119 & 0.8110 & 0.8096 & 0.8098 & 0.8086 & 0.8102 & 0.8061 \\ \cline{3-12}
& & Quarterly & 0.8443 & 0.8441 & 0.8445 & 0.8448 & 0.8448 & 0.8445 & 0.8450 & 0.8424 & 0.8417 \\ \cline{3-12}
& & Monthly & 0.8539 & 0.8552 & 0.8581 & 0.8534 & 0.8551 & 0.8550 & 0.8525 & 0.8562 & 0.8508 \\ \hline
\end{tabular}
\caption{The performance of the CNN and the RNN models, one model is built for each frequency}
\label{owa}
\end{sidewaystable}

The proposed models have very similar performance as FFORMA for daily series.
They do not perform as well for series with much fewer samples, %particularly
i.e., for weekly (359 series) and hourly (414 series) data.
In these cases, a simple linear model works better as it is less prone to overfitting.  Once again, we observe that a large sample size is crucial for NN models to work well for time series forecasting.

In our experiments, one training sample is generated for each series, but it is possible to increase the samples using a stretching window scheme. For a given series $X$ of length $l$, the one training sample is produced by first taking out the last $h$ observations and then applying the $M$ base models to the first $l-h$ observations.
The forecasts by base models are used as the features in the NN model and the last $h$ observations are used as the label. To increase the samples, we can apply the base model to the first $l-h-k$ observations where $k$ is a positive integer, and use the $h$ observations immediately after as the labels. %In this manner, the sample size issue may be

%
%\begin{table}
%\begin{tabular}{|c|c|c|c|c|c|c|c|c|c|}\hline
%  & \multicolumn{3}{|c|}{sMAPE} & \multicolumn{3}{|c|}{MASE} & \multicolumn{3}{|c|}{OWA} \\ \hline
%\diagbox{} {}&
%yearly & quarterly & monthly & yearly & quarterly & monthly & yearly & quarterly & monthly \\ \hline
%1st & 13.176 & 9.679 & 12.126 \\ \hline
%2nd & 13.366 & 9.733 & 12.487 \\ \hline
%3rd & 13.528 & 9.796 & 12.639 \\ \hline
%%4th & 13.669 & 9.800 & 12.737 \\ \hline
%\end{tabular}
%\caption{sMAPE of top submissions}
%\label{smape_ranking}
%\end{table}

\subsection{One RNN model for all frequencies}
As mentioned in Section \ref{sec:cnn}, due to the constraint placed by the fully connected layer, independent CNN models have to be built for series with different forecasting horizons. By contrast, it is possible to build one single RNN model for series of all frequencies. The only restriction is that at the training stage, series of different frequencies have to be in different mini-batches, as the sequence lengths (forecasting horizons) are different.

\begin{sidewaystable}
\centering
\begin{tabular}{|c|c|c|c|c|c|c|c|c|c||c|}\hline
%error measure & \multicolumn{3}{|c|}{sMAPE} \\ \hline
 measure &\diagbox{frequency}{instance}  & 1 & 2 & 3 & 4 & 5 & 6 & 7 & 8 & emsemble \\ \hline
 \multirow{7}{*}{sMAPE} & Yearly & 13.4835 & 13.5163 & 13.4990 & 13.4981 & 13.4544 & 13.5877 & 13.4865 & 13.5747 & 13.4419  \\ \cline{2-11}
& Quarterly & 9.7002 & 9.7302 & 9.7294 & 9.7011 & 9.7082 & 9.7142 & 9.7214 & 9.7143 & 9.6610  \\ \cline{2-11}
& Monthly & 12.5797 & 12.4808 & 12.5347 & 12.4990 & 12.6014 & 12.5821 & 12.5759 & 12.5425 & 12.4405  \\ \cline{2-11}
& Weekly & 8.4877 & 8.7098 & 8.6716 & 8.6312 & 8.5704 & 8.4801 & 8.6964 & 8.6404 & 8.4936  \\ \cline{2-11}
& Daily & 3.0959 & 3.0362 & 3.0286 & 3.0815 & 3.0586 & 3.0190 & 3.0491 & 3.0086 & 3.0354  \\ \cline{2-11}
& Hourly & 13.4411 & 13.3996 & 13.6902 & 13.5543 & 13.8240 & 13.4234 & 13.9391 & 13.4686 & 12.9756 \\ \cline{2-11}
& Total & 11.6845 & 11.6499 & 11.6723 & 11.6497 & 11.6905 & 11.7096 & 11.6893 & 11.6880 & 11.5942  \\ \hline
 \multirow{7}{*}{MASE} & Yearly & 3.1064 & 3.1159 & 3.1086 & 3.1090 & 3.0899 & 3.1384 & 3.0987 & 3.1385 & 3.0980 \\ \cline{2-11}
& Quarterly & 1.1081 & 1.1119 & 1.1098 & 1.1069 & 1.1098 & 1.1105 & 1.1097 & 1.1079 & 1.1043 \\ \cline{2-11}
& Monthly & 0.8999 & 0.8928 & 0.8956 & 0.8933 & 0.8976 & 0.8989 & 0.8994 & 0.8967 & 0.8887 \\ \cline{2-11}
& Weekly & 2.5105 & 2.4393 & 2.4114 & 2.3854 & 2.4421 & 2.4160 & 2.4445 & 2.4334 & 2.3772 \\ \cline{2-11}
& Daily & 3.4150 & 3.3094 & 3.2435 & 3.2838 & 3.2911 & 3.2673 & 3.2930 & 3.2025 & 3.2676 \\ \cline{2-11}
& Hourly & 1.4589 & 1.7442 & 1.6219 & 1.4449 & 1.5087 & 1.5389 & 1.5749 & 1.5111 & 1.4236 \\ \cline{2-11}
& Total & 1.5718 & 1.5679 & 1.5637 & 1.5629 & 1.5620 & 1.5730 & 1.5653 & 1.5685 & 1.5567 \\ \hline
 \multirow{7}{*}{OWA}  &Yearly & 0.8034 & 0.8056 & 0.8041 & 0.8042 & 0.8004 & 0.8106 & 0.8025 & 0.8102 & 0.8011 \\ \cline{2-11}
& Quarterly & 0.8445 & 0.8473 & 0.8465 & 0.8442 & 0.8455 & 0.8461 & 0.8461 & 0.8451 & 0.8414 \\ \cline{2-11}
& Monthly & 0.8593 & 0.8525 & 0.8557 & 0.8534 & 0.8590 & 0.8589 & 0.8589 & 0.8565 & 0.8492 \\ \cline{2-11}
& Weekly & 0.9153 & 0.9146 & 0.9075 & 0.9006 & 0.9075 & 0.8978 & 0.9148 & 0.9097 & 0.8916 \\ \cline{2-11}
& Daily & 1.0293 & 1.0033 & 0.9920 & 1.0069 & 1.0042 & 0.9941 & 1.0030 & 0.9825 & 0.9968 \\ \cline{2-11}
& Hourly & 0.6702 & 0.7286 & 0.7110 & 0.6703 & 0.6910 & 0.6864 & 0.7079 & 0.6818 & 0.6501 \\ \cline{2-11}
& Total & 0.8418 & 0.8395 & 0.8392 & 0.8382 & 0.8395 & 0.8430 & 0.8403 & 0.8411 & 0.8345 \\  \hline
\end{tabular}
\caption{The performance of the RNN model for all frequencies}
\label{owa_rnn}
\end{sidewaystable}

The full results of such a RNN model with a state size of nine are shown in Table \ref{owa_rnn}. Interestingly, this single RNN model built for all series is more accurate than individual models built specifically for a particular frequency. It achieves slightly better OWA than FFORMA for quarterly, monthly and daily series, and does considerably worse for hourly series. Overall, the RNN model is marginally more accurate than FFORMA and would rank 2nd in terms of OWA among all the submissions.

\section{Conclusion}
We present a time series forecasting framework \emph{For2For}. In the framework, forecasts produced by standard models are fed to a NN model, which learns how to make final forecast based on forecasts of various sources. This approach can be seen as a combination method, thus in essence the NN model is trained to learn how to combine forecasts made by standard models. The NN model can be either a CNN model with a structure similar to ResNet or a RNN model which makes forecast one at a time.
%The winning submission in the M4 competition used a hybrid model of exponential smoothing and LSTM networks \cite{smyl2020hybrid}.

To evaluate this approach, we test the method on the M4 competition dataset. Both the CNN and the RNN models perform very well for yearly, quarterly and monthly series. When the data sample size is too small, the NN models tend to overfit and do not generalize well. We also build one single RNN model for all frequencies, and it is more accurate than individual models built specifically for a particular frequency.

In the experiments, we don't use any features other than forecasts generated by base models. In practice, it is certainly possible to combine them. Prediction intervals are not discussed in this paper, and could be a topic of future investigation.

\section*{Acknowledgement}
The work of Ying Feng is supported by XJTLU Research Development Fund 18-02-27.

\end{document}